\documentclass[lettersize,journal]{IEEEtran}
\usepackage{amsmath,amsfonts}
\usepackage{algorithmic}
\usepackage{algorithm}
\usepackage{array}
\usepackage[caption=false,font=normalsize,labelfont=sf,textfont=sf]{subfig}
\usepackage{textcomp}
\usepackage{stfloats}
\usepackage{url}
\usepackage{verbatim}
\usepackage{graphicx}
\usepackage{cite}
\begin{document}

% ------------------------------------ Title ------------------------------------ %

\title{Face-Trace: Open-Set Attribution and Progressive Discovery of Synthetic Face Generators}

\author{Alessia~Infantino,
        Claudio~Schiavella~\IEEEmembership{Student~Member~IEEE,}
        and~Irene~Amerini~\IEEEmembership{Member~IEEE}%
\thanks{All authors are with the Department of Computer, Control and Management Engineering (DIAG), Sapienza University of Rome, 00185 Rome, Italy (e-mail: infantino.1922069@studenti.uniroma1.it; schiavella@diag.uniroma1.it; amerini@diag.uniroma1.it).}}

% The paper headers
\markboth{Preprint. Under review.}{Infantino \MakeLowercase{\textit{et al.}}: Face-Trace: Open-Set Attribution and Progressive Discovery of Synthetic Face Generators}

\maketitle

% ------------------------------------ Abstract ------------------------------------ %
\begin{abstract}
Recent advances in generative Artificial Intelligence have made synthetic face images increasingly realistic, creating new challenges for multimedia forensics. Source attribution methods should identify the generator of an image when the source is known, but also handle samples produced by unseen models. Most existing approaches, however, address synthetic face attribution in a closed-set setting, assuming that test samples can only originate from generators observed during training. This assumption does not hold in real-world scenarios, where new generators continuously appear and detecting an image as unknown is not sufficient, since rejected samples should also be organized according to their underlying sources. We introduce Face-Trace, a pipeline for open-set synthetic face source attribution that combines known generator classification, energy-based rejection, and unknown generator discovery. A classifier trained on frozen I-JEPA embeddings attributes known generators, while rejected samples are represented by combining projected I-JEPA features with complementary forensic traces and grouped to identify coherent sets of samples produced by unknown generators. We also extend the discovery stage to an incremental scenario, where rejected samples arrive over time. Experiments on the WILD dataset show 96.73\% closed-set attribution accuracy, while rejection reaches 71.25\% balanced accuracy and rejected samples are clustered into meaningful unknown-generator groups, with an Adjusted Rand Index of 0.81, a Normalized Mutual Information of 0.90, and an overall purity of 87.74\%. In the incremental setting, the discovered generator space is progressively extended while maintaining a final purity of 99.23\%, and cross-dataset experiments suggest that the pipeline can operate beyond the original data distribution.
\end{abstract}


% ------------------------------------ Introduction ------------------------------------ %

\section{Introduction}
The rapid advancement of Artificial Intelligence and the diffusion of generative models has made it possible to create synthetic images with a very high level of realism~\cite{nightingale2022aisynthesized}. In particular, modern text-to-image models, GAN-based models, and diffusion models can generate human faces that look realistic and are difficult to distinguish from real ones~\cite{fernando2025facedeepfakes}. While these technologies support legitimate applications in cinema, digital entertainment, advertising, and creative production, they also raise new challenges for multimedia forensics, where understanding the origin of a visual content is becoming increasingly important~\cite{amerini2024deepfakeforensics,amerini2026ff4all}.

In this context, it is important to distinguish between synthetic image detection and source attribution~\cite{khoo2021deepfake}. Synthetic image detection aims to understand whether an image is real or generated. Source attribution, instead, aims to identify which generative model produced the image. In this work, we focus on synthetic face source attribution, where the goal is to assign the image to its generator.

Most existing attribution methods work in a closed-set scenario~\cite{bongini2025wild}. In this setting, the generators used during testing are the same generators seen during training. Therefore, each test image is assumed to belong to one of the known generator classes. New generative models are continuously released, and images may come from sources that were not available when the attribution system was trained. As a result, a closed-set classifier can produce unreliable predictions when it receives samples generated by unseen models. 
Open-set attribution addresses this limitation by considering both In-Distribution (ID) samples, produced by known generators (i.e. generators included during training), and Out-Of-Distribution (OOD) samples, produced by unseen generators~\cite{salehi2021unified}. In this setting, the attribution system should not only classify ID generators correctly, but also detect OOD samples. Rejection scores, such as confidence-based or energy-based scores, can be used for this purpose~\cite{hendrycks2016baseline}.
However, rejection only separates known from unknown samples and it does not explain how the unknown samples are related to each other. For this reason, open-set source attribution should also include a discovery stage where rejected OOD samples are organized into coherent groups that can help distinguish different unknown generators~\cite{azizpour2025autonomous}.

In this paper, we introduce \textit{Face-Trace}, a pipeline for open-set synthetic face source attribution. \textit{Face-Trace} first attributes samples produced by known generators and uses an energy-based score to reject samples that are likely to originate from unseen sources. Rejected samples are then grouped in the embedding space to discover coherent sets of unknown generators. 
We also extend this setting to an incremental discovery scenario. This is motivated by the fact that in realistic conditions rejected samples are not available all at once, but arrive progressively over time. The system starts from an initial set of discovered unknown-generator groups, each associated with a pseudo-label, i.e., an identifier that distinguishes the discovered source without requiring its true generator identity. New rejected samples that are close to an existing group inherit its pseudo-label, while unmatched samples are stored in a buffer. When enough buffered samples are collected, they are clustered to identify possible new generators, and sufficiently reliable clusters receive new pseudo-labels and are added to the unknown-generator space. This allows the system to progressively expand the set of discovered sources.
Unlike continual-learning approaches, \textit{Face-Trace} does not retrain the backbone, the known-generator classifier, or the rejection module when a new source is discovered, it updates only the unknown-generator space.

Since all images considered in this work are synthetic faces, they share a similar semantic structure. This controlled setting allows us to evaluate whether a robust and transferable representation can still preserve differences related to the generative source. For this reason, we use a frozen I-JEPA encoder~\cite{assran2023ijepa} as visual backbone and combine its representation with forensic traces designed to capture source-specific artifacts, with the goal of supporting known-generator attribution, OOD rejection, and unknown-generator discovery.

The main contributions of this work are:
\begin{itemize}
    \item We introduce \textit{Face-Trace}, an open-set synthetic face attribution framework that combines energy-based rejection with embedding clustering, moving from unknown-generator detection to unknown-generator discovery.
    Unlike prior self-adapting attribution systems~\cite{azizpour2025autonomous}, our discovery stage operates on a frozen space with fused representation and does not require retraining the model as new generators are discovered.

    \item We formulate unknown-generator discovery in a non-transductive and incremental setting, where samples from unseen generators are not available during training and are processed progressively as they arrive.

    \item We study the use of I-JEPA representations combined with forensic descriptions to cluster rejected samples into generator-specific groups. To the best of our knowledge, this is the first work to investigate frozen I-JEPA representations for synthetic image source attribution and unknown-generator discovery.

    \item We evaluate the method on the WILD dataset~\cite{bongini2025wild} and on external datasets, analyzing closed-set attribution, OOD rejection, unknown-generator clustering, incremental discovery, cross-dataset transfer, and robustness to post-processing.

\end{itemize}

\section{Related Work}
\subsection{Out-of-Distribution Detection}
OOD detection addresses the problem of identifying test samples that do not follow the same distribution as the data used to train a model~\cite{yang2024generalizedood}. Several rejection scores have been proposed to identify OOD samples. Early approaches rely on the Maximum Softmax Probability, which uses the maximum predicted class probability as a confidence score and samples with low confidence are rejected as OOD~\cite{hendrycks2016baseline}. Other methods exploit information from the logits or gradients, such as MaxLogit~\cite{hendrycks2019anomalysegmentation}, GradNorm~\cite{huang2021gradients}, or energy scores that assign lower energy to In-Distribution (ID) samples and higher energy to OOD samples, based on model logits~\cite{liu2020energy}. More recently, Generalized Entropy has been proposed as a softmax-based score designed to amplify small deviations from confident, near one-hot predictions~\cite{liu2023gen}. In standard OOD detection, however, a sample is either accepted or rejected, whereas different OOD samples may originate from different unseen classes or domains. For this reason, an additional discovery stage is needed to group samples into separate unknown categories.

\subsection{Open-Set Synthetic Image Attribution}
Most existing attribution methods assume a closed-set scenario. However, synthetic images are produced by a rapidly increasing number of generators, meaning that the appearance of new generators naturally defines an out-of-distribution condition for the proposed systems~\cite{cai2026towards}. As a consequence, a closed-set classifier may force samples from unseen generators into one of the known classes, resulting in incorrect attributions~\cite{yang2023pose}.
Some works attempt to improve generalization by regularizing the learned representation. For example, DNA-Det~\cite{yang2022dnadet} regularizes the representation through transformation-based pretraining and patchwise contrastive learning which, combined with post-hoc rejection strategies, improves robustness to unseen samples.
Other methods explicitly address the separation between known and unknown sources. For instance, RepMix~\cite{bui2022repmix} uses a training objective that pushes the model to assign high confidence to embeddings from known classes and low scores to embeddings from unknown ones; a threshold is then applied to distinguish known from unknown data. Similarly, POSE~\cite{yang2023pose} expands the feature space of known classes by generating synthetic variations that simulate unknown model traces, enabling a more accurate estimation of the boundary between known and unknown classes. More recently, BOSC~\cite{wang2025bosc} introduces a backdoor-based classifier with a rejection mechanism for open-set synthetic image attribution. However, rejected samples are treated as belonging to an unknown class and are not further grouped to discover their source generators. Facial deepfake detection has also been studied in the open-set setting, where samples generated by unknown generators are flagged as unknown~\cite{bahavan2026deepfake}; this approach uses a weighted supervised contrastive learning strategy to obtain a more discriminative latent space.
These works, however, are still formulated as rejection problems and do not further organize samples into meaningful novel classes.

\subsection{Open-Set Generalization with Foundation Models} 
Facing open-set scenarios, recent works have explored more general representations to improve robustness. In these approaches, foundation models such as CLIP~\cite{radford2021clip} and DINOv3~\cite{simeoni2025dinov3} are often considered, since they are large pretrained encoders that learn transferable visual representations without being trained specifically for synthetic image forensics. In the same direction, self-supervised architectures such as I-JEPA~\cite{assran2023ijepa} have also become relevant. I-JEPA learns representations by predicting masked parts of the input in the latent space, and this encourages the model to capture high-level and transferable visual structures.
In parallel, contrastive learning aims to structure the embedding space by bringing similar samples closer while pushing dissimilar ones apart, which can help obtain more discriminative representations. For example, some works employ supervised contrastive learning to obtain discriminative embeddings and then apply a k-NN classifier within a few-shot learning paradigm to enhance adaptability to novel generative models~\cite{uruena2025supervised}. Other works use image-text contrastive learning to learn generator-agnostic representations for detecting synthetic images from unseen generators~\cite{wu2026generalizable}. Similarly, a pre-trained CLIP encoder has been used to generate image-text embeddings for detection and attribution in both closed-set and open-set scenarios through confidence-based rejection~\cite{sha2023defake}.
Intermediate visual features from CLIP and DINOv2 have also been used for synthetic image attribution, combined with classifiers such as logistic regression or k-NN for known-generator classification and post-hoc rejection scores for samples from unseen sources~\cite{cioni2025clip}.
All of these methods avoid relying solely on specific forensic artifacts and instead leverage richer contextual information to support attribution, suggesting that a robust embedding space is crucial for both attribution and open-set generalization.

\subsection{Novel Category Discovery}
Novel Category Discovery (NCD) studies how to discover categories that are not labeled during training. In the standard NCD setting, the training data is composed of a labeled set from known classes and an unlabeled set from disjoint novel classes; the goal is to partition the unlabeled samples into coherent novel categories~\cite{han2019deeptransfer,troisemaine2023ncd}. 
This differs from purely unsupervised clustering, since the labeled known classes provide prior knowledge about what constitutes a meaningful category and help learn a more suitable representation or similarity function~\cite{troisemaine2023ncd}. For example, Deep Transfer Clustering~\cite{han2019deeptransfer} learns a representation from known classes and then clusters unlabeled samples from novel classes.

Generalized Category Discovery (GCD) extends NCD to a more realistic setting, where the unlabeled set may contain both known and novel classes. GCD~\cite{vaze2022generalized} addresses this problem through contrastive representation learning followed by semi-supervised clustering. Similarly, SimGCD~\cite{wen2023parametric} learns a parametric classifier by exploiting labeled known classes and unlabeled mixed data, using self-distillation and entropy regularization to recognize seen categories and discover novel ones. More recently, ProtoGCD~\cite{ma2025protogcd} learns shared prototypes for known and novel classes through adaptive pseudo-labeling. In the forensic domain, category discovery is closely related to the discovery of unknown manipulation or generation sources. CAL~\cite{zheng2025openworld} studies open-world deepfake attribution, where the goal is to attribute both known and novel face forgery types. 

However, many NCD and GCD methods assume a transductive setting, where the unlabeled samples to be grouped are already available during training or adaptation. This allows the method to exploit the target data distribution before inference. In contrast, in a non-transductive setting, novel samples are not seen during training and must be processed only when they appear at test time. On-the-fly Category Discovery (OCD)~\cite{du2023onthefly} addresses this setting using hash-based category descriptors and sign-magnitude disentanglement to group seen and unseen categories. Our problem follows a similar formulation, since images from unknown generators become available only after training and must be grouped according to their source.

\subsection{Unknown Generator Discovery and Clustering}
Building on the idea of NCD and GCD, unknown generator discovery focuses on grouping unknown samples according to their source generator. Previous work has shown that embeddings extracted from self-supervised encoders can often be clustered according to semantic classes~\cite{lowe2024zeroshot}. Similarly, clustering-based OOD detection methods rely on the assumption that ID samples form compact groups in the embedding space, while OOD samples lie far from these clusters~\cite{sinhamahapatra2022cluster}. However, these methods mainly group samples according to their semantic content or use clustering only to distinguish ID from OOD, rather than to identify previously unseen generators. Related fingerprint-based work such as AdaParse~\cite{zheng2026adaparse} learns image-specific fingerprints to infer architectural and training hyperparameters of visual generative models, but does not address open-set rejection or unknown-generator discovery. Only a limited number of works explicitly investigate clustering for unknown source discovery. An iterative pipeline has been proposed for discovering and attributing GAN images, where OOD samples are clustered, merged, refined, and used to assign pseudo-labels to unseen sources~\cite{girish2021openworldgan}. More recently, Forensic Self-Descriptions (FSD) have been introduced to model image forensic microstructures and support zero-shot detection, open-set source attribution, and unsupervised clustering of image sources~\cite{nguyen2025forensic}. Incremental learning has also been applied to AI-generated image detection~\cite{tang2025extensible}. In this approach, the detector is updated whenever labeled images from a new generator become available, while knowledge distillation and domain alignment are used to preserve performance on previously seen generators. Closer to our setting, another recent work explores autonomous, self-adapting systems in which FSD are used to learn a more separated embedding space: rejected samples are stored in a buffer, clustered to discover new generators, and then used to update the model~\cite{azizpour2025autonomous}.

Our work is related to these FSD-based attribution approaches, but differs in several important aspects. First, instead of relying only on forensic descriptions, we combine FSD with a projected I-JEPA representation. Second, the attribution and rejection modules are kept fixed after training: newly discovered clusters are used to update the reliable unknown-generator space, but they are not used to retrain the known-generator classifier. Finally, we evaluate the discovery stage in both offline and incremental settings. In the incremental scenario, rejected samples arriving over time are either assigned to previously discovered sources or stored as evidence for emerging generators. We also assess the method on cross-dataset streams and under post-processing.

\section{Problem Formulation}
\subsection{Closed-Set Attribution}
Let $\mathcal{X} = \mathbb{R}^{C \times H \times W}$ denote the image space and let $\mathcal{Y}_{K} = \{1, \dots, K\}$ be the set of known generator classes available during training. Given a training set

\begin{equation}
    \mathcal{D}_{\mathrm{train}} = \{(x_i, y_i)\}_{i=1}^{N},
\end{equation}

where $x_i \in \mathcal{X}$ is a synthetic face image and $y_i \in \mathcal{Y}_{K}$ is its source-generator label, the closed-set source attribution task consists in learning a classifier $f: \mathcal{X} \rightarrow \mathcal{Y}_{K}$ that assigns each image to one of the known generators.

\subsection{Open-Set Attribution}
The closed-set formulation assumes that every test sample is produced by one of the $K$ known generators. This assumption is unrealistic, since at test time we may encounter images generated by sources unseen during training. We denote the set of unknown generators as

\begin{equation}
    \mathcal{Y}_{U} = \{K+1, \dots, K+U\},
\end{equation}

with $\mathcal{Y}_{U} \cap \mathcal{Y}_{K} = \emptyset$, and we assume that samples from $\mathcal{Y}_{U}$ are not available during training. At test time, a sample may originate from either $\mathcal{Y}_{K}$ or $\mathcal{Y}_{U}$, i.e.,

\begin{equation}
    \mathcal{D}_{\mathrm{test}} = \{x_j\}_{j=1}^{M}, \qquad
    y_j \in \mathcal{Y}_{K} \cup \mathcal{Y}_{U}.
\end{equation}

To handle this setting, we equip the system with a rejection score
$s: \mathcal{X} \rightarrow \mathbb{R}$ and a threshold $\tau$. The open-set classifier attributes a sample to a known generator when its score is below the threshold, and rejects it as unknown otherwise:
\begin{equation}
    f_{\mathrm{os}}(x) =
    \begin{cases}
        f(x), & \text{if } s(x) \leq \tau,\\
        u, & \text{if } s(x) > \tau,
    \end{cases}
\end{equation}

where $u \notin \mathcal{Y}_{K}$ denotes the rejection, i.e., an unknown outcome. The set of rejected samples is then

\begin{equation}
    \mathcal{R} = \{\, x \in \mathcal{D}_{\mathrm{test}} : s(x) > \tau \,\}.
\end{equation}

\subsection{Unknown Generator Discovery}
Rejecting a sample as unknown is not sufficient, since $\mathcal{R}$ may contain images produced by several distinct unknown generators. Moreover, because rejection is imperfect, $\mathcal{R}$ may also include a fraction of samples from known generators that are erroneously rejected. The discovery stage should therefore organize the rejected samples into groups, each ideally corresponding to a single unseen generator, while remaining robust to this residual contamination from known sources. Discovery is performed on a dedicated representation $\psi: \mathcal{X} \rightarrow \mathbb{R}^{d_\psi}$, which may differ from the classification representation space.

We first consider an offline setting, in which a pool of rejected samples is available at once and is partitioned into a set of clusters

\begin{equation}
    \mathcal{S} = \{S_1, \dots, S_{Q}\},
\end{equation}

where each cluster $S_q$ ideally corresponds to a distinct unknown generator. The number of unknown generators $Q$ is estimated and not assumed to be known in advance; in the ideal case $Q = U$ and the clusters recover $\mathcal{Y}_{U}$ up to a permutation of the labels. This offline clustering also serves to initialize the incremental setting described below, by producing the initial set of reliable clusters $\mathcal{S}_{0}$.

In realistic conditions, rejected samples do not arrive all at once but
progressively over time. Let 

\begin{equation}
    \mathcal{S}_{t} = \{S_1, \dots, S_{Q_t}\}
\end{equation}

denote the reliable unknown space, i.e., the set of clusters that have been validated and promoted, available at step $t$, where each previously discovered cluster $S_q$ is associated with a pseudo-label $\tilde{y}_q \in \tilde{\mathcal{Y}}_{t}$ with $\tilde{\mathcal{Y}}_{t} \cap \mathcal{Y}_{K} = \emptyset$. Pseudo-labels are assigned incrementally as new generators are discovered, and they do not necessarily correspond to the true unknown identities in $\mathcal{Y}_{U}$: they only serve to consistently distinguish one discovered source from another. The space is initialized as $\mathcal{S}_{0}$ from the offline clustering and is then expanded over time.

\begin{figure*}[!t]
    \centering
    \includegraphics[width=\textwidth]{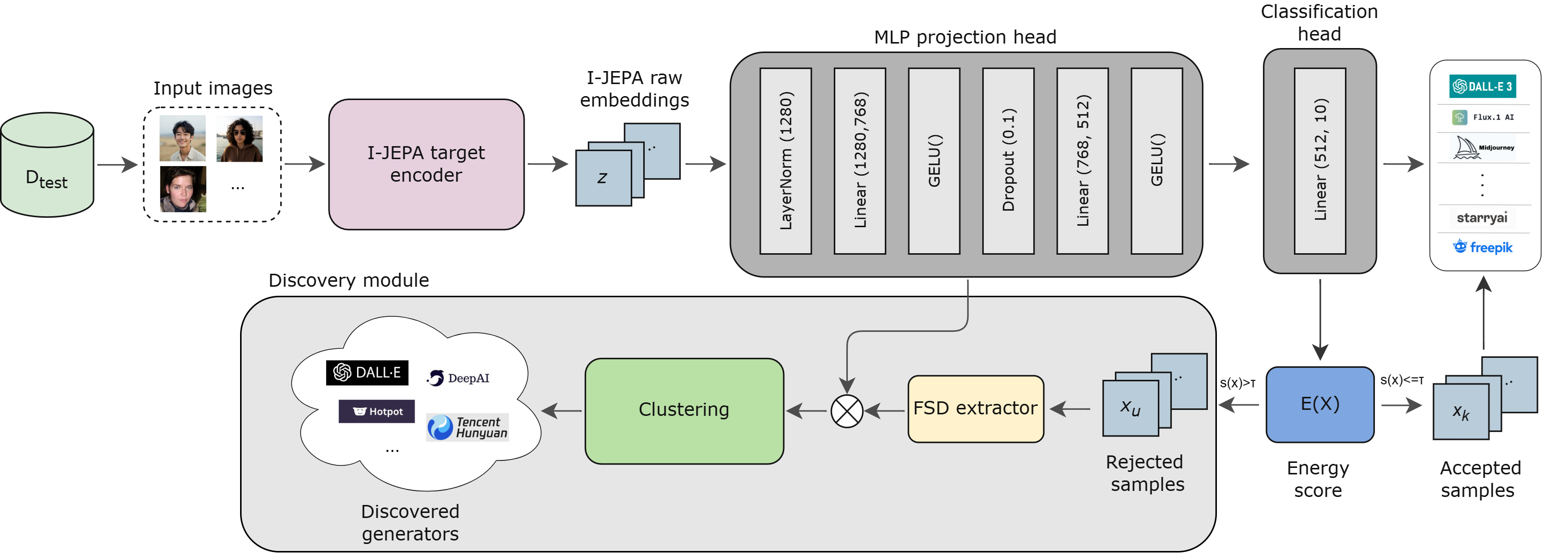}
    \caption{Overview of the Face-Trace offline pipeline. Given an input image, the frozen I-JEPA target encoder extracts a raw embedding, which is processed by the MLP projection head and the classification head to produce logits over the known generators. The logits are used to compute an energy score for open-set rejection: samples with low energy are accepted and attributed to one of the known generators, while high-energy samples are rejected as unknown. For rejected samples, the pipeline combines the projected I-JEPA representation with the corresponding FSD. The resulting fused representations are then clustered with HDBSCAN to discover and organize samples from unknown generators.}
    \label{fig:pipeline}
\end{figure*}

When a rejected sample $x$ arrives at step $t$, the system first attempts to match it to an existing reliable cluster by computing a distance $d(\psi(x), S_q)$ in the discovery space and selecting the nearest one, $q^{*} = \arg\min_{q \in \{1, \dots, Q_t\}} d(\psi(x), S_q)$. The sample is assigned to $S_{q^{*}}$, and inherits its pseudo-label $\tilde{y}_{q^{*}}$, when it falls within the matching radius of that cluster; otherwise it is stored in a buffer for later discovery:

\begin{equation}
    \begin{cases}
    x \rightarrow S_{q^{*}}, & \text{if } d(\psi(x), S_{q^{*}}) \leq \delta_{q^{*}},\\
    x \rightarrow \mathcal{B}_{t}, & \text{otherwise,}
    \end{cases}
\end{equation}

where $\delta_{q}$ is a per-cluster matching radius that adapts to the spread of each reliable cluster $S_q$.

When the buffer reaches a predefined size $B$, a clustering operator is applied to the buffered embeddings to produce a set of candidate clusters

\begin{equation}
    \widehat{\mathcal{S}}_{t} = \{\widehat{S}_1, \dots, \widehat{S}_{\widehat{Q}_t}\},
\end{equation}

where each $\widehat{S}_r$ represents a possible new unknown generator. A candidate cluster is eligible for promotion only when it is sufficiently supported and compact, i.e.,

\begin{equation}
     \mathcal{P}_{t} = \{\, \widehat{S}_r \in \widehat{\mathcal{S}}_{t} :|\widehat{S}_r| \geq n_{\min} \;\wedge\; \mathrm{disp}(\widehat{S}_r) \leq \rho_{\max} \,\},
\end{equation}

where $\mathrm{disp}(\cdot)$ measures the dispersion of a cluster in the embedding space, $n_{\min}$ is the minimum support, and $\rho_{\max}$ is the maximum admissible dispersion. A cluster therefore enters $\mathcal{P}_{t}$ only when it is both sufficiently populated and compact. 
Each selected cluster receives a new pseudo-label and is added to the reliable unknown space, which is updated as

\begin{equation}
    \mathcal{S}_{t+1} = \mathcal{S}_{t} \cup \mathcal{P}_{t}.
\end{equation}

At each subsequent step of the stream, the updated space $\mathcal{S}_{t+1}$ is used for matching, i.e., to compare each newly arriving rejected sample against the discovered clusters and assign it to an existing source when compatible. In this way, the system progressively builds a set of discovered unknown sources as new samples arrive.

\section{Methodology}
\subsection{Feature Extraction}
The \textit{Face-Trace} pipeline, illustrated in Fig.~\ref{fig:pipeline}, starts by extracting embeddings from the input images using a frozen pre-trained I-JEPA target encoder~\cite{assran2023ijepa}. I-JEPA belongs to the family of visual foundation models, such as CLIP~\cite{radford2021clip} and DINOv3~\cite{simeoni2025dinov3}, which can be used as general-purpose feature extractors. While CLIP learns image representations through image-text alignment and DINOv3 learns self-supervised features through view consistency, I-JEPA learns by predicting missing image information directly in the latent space.

The choice of I-JEPA is motivated by the structure of the problem: all images in our experiments are synthetic faces and therefore share similar semantic content. In this controlled semantic setting, we evaluate whether a high-level visual representation can still preserve differences related to the generative source and support source attribution, open-set rejection, and unknown generator discovery.

\subsection{Known Generator Classification}
After extracting the I-JEPA embeddings, we train a classifier on the extracted feature vectors to attribute the source images to the corresponding generator classes. 

We use a Multi-Layer Perceptron to keep the attribution module lightweight and easy to train. It is composed of a projection head followed by a linear classifier. The projection head first applies Layer Normalization to the 1280-dimensional I-JEPA embedding, and then maps it to a 512-dimensional representation through two fully connected layers with GELU activations. A dropout layer is used after the first hidden layer to reduce overfitting. The resulting 512-dimensional representation is then fed to the linear classifier where logits are converted into class probabilities through a softmax function and are then used for closed-set attribution.
In the open-set scenario, the logits are also used to compute the energy score, which measures whether the sample is compatible with the known generator classes or should be rejected as unknown.

\subsection{OOD Rejection Module}
After the classifier produces the logits for an input image, the open-set rejection module decides whether the sample should be accepted or rejected as unknown using an energy-based rejection score.

Given the logits produced by the classifier, the energy score is computed as:
\begin{equation}
E(x) = -T \log \sum_{k=1}^{K} \exp \left( \frac{l_k(x)}{T} \right),
\end{equation}
where $l_k(x)$ is the logit associated with the $k$-th known generator class, $K$ is the number of known classes, and $T$ is the temperature parameter~\cite{liu2020energy}.
The threshold $\tau$ is selected on the validation set using only known-generator samples, by fixing the false positive rate at 5\%. No samples from unseen generators are used for threshold calibration. A sample is accepted when $E(x) \leq \tau$ and rejected otherwise.
Images from known generators usually produce lower energy values, because the classifier is more confident about known classes, while images from unseen generators usually produce higher energy values.
Finally, accepted samples are assigned to one of the known generator classes, while rejected samples go to the generator discovery module.

\subsection{Novel Generator Discovery}
For novel generator discovery, source-specific forensic traces are needed to complement the representation learned for known-generator classification. We therefore combine the projected I-JEPA embedding with FSD features~\cite{nguyen2025forensic}, which capture complementary forensic information.

The frozen I-JEPA encoder produces a 1280-dimensional embedding, which is mapped by the projection head into a 512-dimensional representation. The FSD extractor produces a 960-dimensional descriptor of source-specific forensic microstructures derived from the image residuals. Since the two feature types have different dimensionalities, both branches are reduced to 64 dimensions and then concatenated, resulting in a final 128-dimensional embedding.
We consider a non-transductive open-set attribution setting: during training, the model has access only to labeled samples from known generators. At test time, samples from unknown generators are first rejected by the OOD module and then processed by the discovery stage.

In the incremental setting, an initial batch of rejected samples is clustered to initialize a reliable unknown-generator space, defined as a set of validated clusters representing discovered generators. For each subsequent rejected sample, the system first tries to match it to one of the reliable clusters already present in this space.  To do this, we compute the Mahalanobis distance~\cite{mahalanobis} between the sample embedding and each reliable cluster. Given the embedding $\psi(x)$ of a rejected sample and a reliable cluster $S_q$, the Mahalanobis distance is defined as

\begin{equation}
d_M(\psi(x),S_q)
=
\sqrt{
(\psi(x)-\boldsymbol{\mu}_q)^{T}
\boldsymbol{\Sigma}_q^{-1}
(\psi(x)-\boldsymbol{\mu}_q)
},
\end{equation}

where $\boldsymbol{\mu}_q$ and $\boldsymbol{\Sigma}_q$ are the centroid and covariance matrix of cluster $S_q$, respectively. Unlike Euclidean distance, the Mahalanobis distance accounts for the spread and direction of the cluster in the embedding space.
The sample is then compared with all reliable clusters and assigned to the closest one if its distance is within a specific matching radius. In this case, it inherits the cluster pseudo-label; otherwise, it is stored in a buffer for later clustering.

\begin{figure}[!t]
\centering
\includegraphics[width=2.5in]{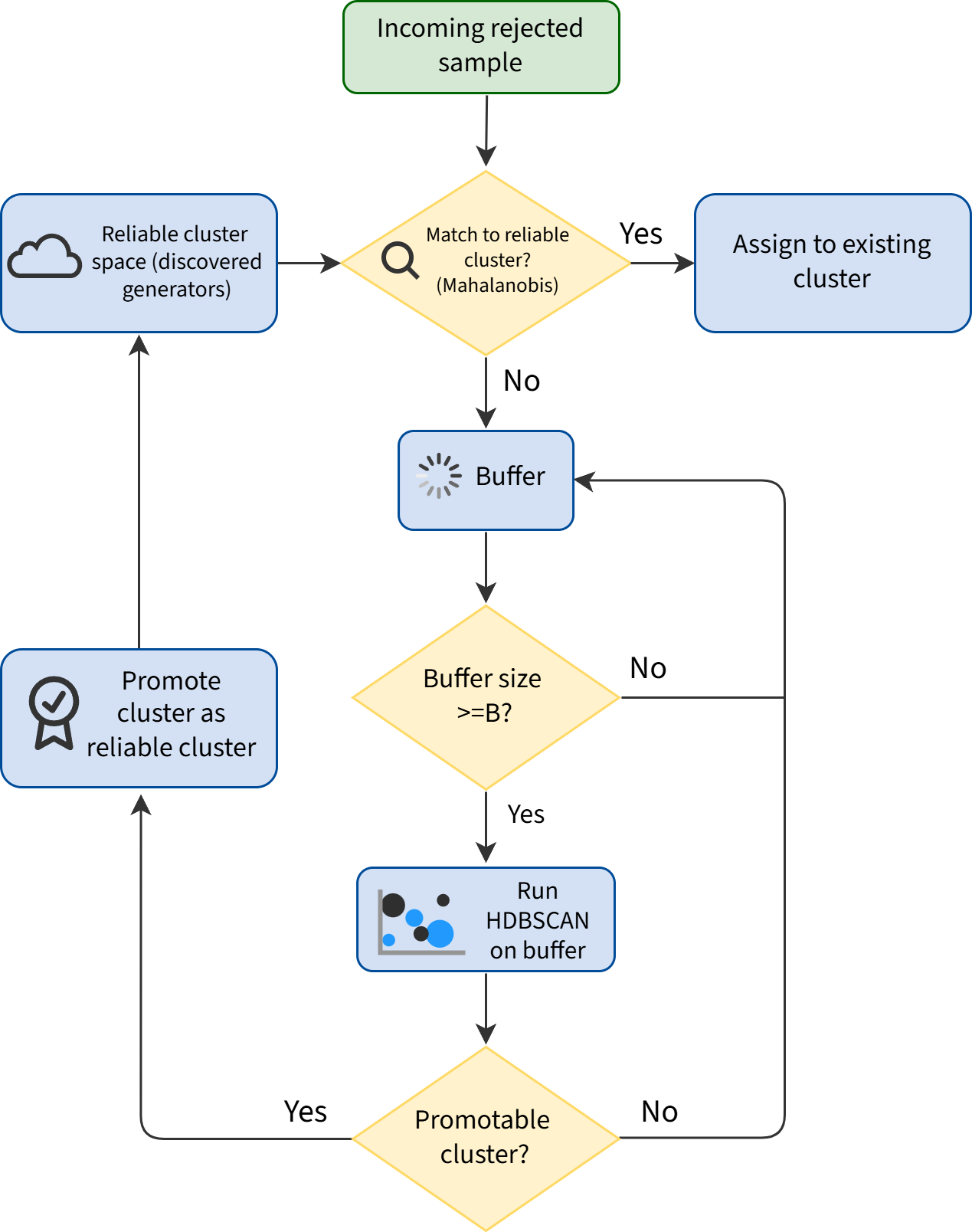}
\caption{Overview of the incremental discovery logic of Face-Trace.}
\label{incremental_img}
\end{figure}

When the buffer reaches a predefined size, HDBSCAN~\cite{campello2013hdbscan} is applied to the buffered embeddings to identify new candidate clusters. Unlike K-Means or Gaussian-mixture models (GMM), HDBSCAN does not require the number of clusters to be specified in advance and can label ambiguous samples as noise.

Not all clusters found in the buffer are promoted. A candidate cluster is promoted only if it contains enough samples and has sufficient cohesion, otherwise its samples remain in the buffer (Fig.~\ref{incremental_img}). Samples that remain unassigned after a predefined number of buffer-clustering steps are removed to prevent noise accumulation.

\subsection{Pseudo-Label Assignment}
Once a candidate cluster is promoted, it receives a pseudo-label and is added to the reliable unknown-generator space as a newly discovered generator. Subsequent rejected samples can then be matched against this cluster.
If a new rejected sample is compatible with a promoted cluster, it receives the corresponding pseudo-label instead of being inserted into the buffer again.
In this way, the system progressively builds a reliable space of discovered unknown generators, which is used to assign subsequent rejected samples to previously identified sources or detect new ones.

\section{Evaluation Metrics}
We evaluate \textit{Face-Trace} at different stages: closed-set attribution, OOD rejection, unknown generator clustering, and incremental discovery.

\subsection{Closed-Set Attribution and OOD Rejection Metrics}
Closed-set attribution is evaluated using accuracy, AUC, and confusion matrices. For OOD rejection, we report the True Positive Rate (TPR), corresponding to the fraction of OOD samples correctly rejected, the True Negative Rate (TNR), corresponding to the fraction of ID samples correctly accepted, and the balanced rejection accuracy
\begin{equation}
    \mathrm{BAcc}
    = \frac{\mathrm{TPR}+\mathrm{TNR}}{2}.
\end{equation}
We also report the area under the ROC curve (AUROC).

End-to-end open-set attribution is evaluated using the Open Set Classification Rate (OSCR). For a rejection threshold $\tau$, the Correct Classification Rate (CCR) and False Positive Rate (FPR) are defined as
\begin{equation}
\begin{aligned}
\mathrm{CCR}(\tau)
&=
\frac{
|\{x \in \mathcal{D}_{ID}: s(x)\leq\tau \wedge f(x)=y\}|
}{
|\mathcal{D}_{ID}|
},\\
\mathrm{FPR}(\tau)
&=
\frac{
|\{x \in \mathcal{D}_{OOD}: s(x)\leq\tau\}|
}{
|\mathcal{D}_{OOD}|
}.
\end{aligned}
\end{equation}
We report the area under the OSCR curve (AUOSCR) and CCR at 5\% FPR.

\begin{table*}[!t]
\caption{Dataset composition and role in the experimental protocol. "Seen" indicates a generator already present in WILD (as a known or previously discovered source), used to test cross-dataset consistency; "Unseen" indicates a generator never encountered before.}
\label{tab:dataset_setup}
\centering
\renewcommand{\arraystretch}{1.2}
\begin{tabular}{p{1.8cm} p{2.1cm} p{3.2cm} p{1.3cm} p{7.2cm}}
\hline
Dataset & Split / Source & Role & Samples & Generators \\
\hline
WILD~\cite{bongini2025wild} & Closed-set & Known generators & 10000 &
Adobe Firefly, DALL-E 3, Flux.1, Flux.1.1 Pro, Freepik, Leonardo AI, Midjourney, Stable Diffusion 3.5, Stable Diffusion XL, Starry AI \\

WILD~\cite{bongini2025wild} & Open-set & Unknown generators & 10000 &
DALL-E 1, Deep AI, Hotpot AI, Nvidia Sana PAG, Stable Cascade, Stable Diffusion Attend and Excite, StyleGAN, StyleGAN2, StyleGAN3, Tencent Hunyuan \\

WILD~\cite{bongini2025wild} & Post-processed closed-set & Robustness evaluation & 3000 per-step &
Adobe Firefly, DALL-E 3, Flux.1, Flux.1.1 Pro, Freepik, Leonardo AI, Midjourney, Stable Diffusion 3.5, Stable Diffusion XL, Starry AI \\

WILD~\cite{bongini2025wild} & Post-processed open-set & Robustness evaluation & 5000 per-step &
DALL-E 1, Deep AI, Hotpot AI, Nvidia Sana PAG, Stable Cascade, Stable Diffusion Attend and Excite, StyleGAN, StyleGAN2, StyleGAN3, Tencent Hunyuan \\

SFHQ-T2I~\cite{beniaguev2024sfhq} & External  Open-set & Seen known generator & 1000 &
DALL-E 3 \\

SoFake~\cite{huang2025sofake} & External Open-set & Seen unknown generator & 1000 &
StyleGAN2 \\

AI-Face~\cite{lin2025aiface} & External Open-set & Unseen unknown generators & 2000 &
AttGAN, Latent Diffusion \\
\hline
\end{tabular}
\end{table*}

\subsection{Unknown Generator Clustering Metrics}
Clustering quality is evaluated using Adjusted Rand Index (ARI), Normalized Mutual Information (NMI), and purity, which respectively measure agreement with the true generator partition, shared information between predicted and true labels, and cluster homogeneity. Unless otherwise specified, these metrics are computed only on samples not labelled as noise by HDBSCAN.

\subsection{Incremental Discovery Metrics}
For the incremental discovery stage, we evaluate both the matching to already discovered unknown generators and the discovery of novel generators. 

For samples from previously discovered generators, the Direct Match Rate (DMR) measures the fraction assigned immediately to an existing cluster. The Final Assignment Rate (FAR) includes assignments obtained both directly and after buffering, while the Still Buffered Rate (SBR) measures the fraction of buffered samples that remain unassigned at the end of the stream. We additionally report the accuracy of the assigned pseudo-labels, the number and purity of promoted clusters, and the final purity of the reliable unknown space.

\section{Experimental Setup}
\subsection{Datasets}
We evaluate \textit{Face-Trace} on the WILD dataset~\cite{bongini2025wild}, which contains face images generated by 20 different generators. Each generator has 1000 images, for a total of 20000 images. The dataset is further organized in 10 closed-set generators and 10 open-set generators.
In our experiments, the closed-set generators are treated as known classes (ID), whereas the open-set generators are treated as unknown classes (OOD). The closed set is used for training, validation, and closed-set testing, while the open-set is used only at test time to evaluate rejection and novel generator discovery.

WILD also provides post-processed versions of the images by applying one to three randomly selected image transformations, including compression, cropping, resizing, rotation, brightness and contrast changes, blur, grayscale conversion, super-resolution, and JPEG AI compression. 
The post-processed images are grouped by transformation depth: 1-step, 2-step, and 3-step. In this work, these images are used only at test time to evaluate how robust unknown generator clustering is to increasingly modified images.

In addition to the experiments conducted on WILD, we perform a cross-dataset evaluation using external datasets. This setting allows us to test whether the proposed pipeline can handle samples coming from different data distributions, including samples from known generators, generators discovered by our pipeline, and completely unseen generators. All images are resized to $224 \times 224$ and normalized using ImageNet mean and standard deviation.

Table~\ref{tab:dataset_setup} summarizes the datasets used in the experiments, together with their role in the proposed evaluation protocol.

\subsection{Training Setup}
The classification head is trained on the WILD closed-set training split. The closed-set is divided into 5000 training images, 2000 validation images, and 3000 test images. 

During training, the I-JEPA encoder is kept frozen, and the classifier is trained with Cross-Entropy loss (CE) with label smoothing equal to 0.1. We use AdamW with a learning rate of $5 \times 10^{-4}$ and a weight decay of $10^{-3}$. The batch size is 32, the maximum number of epochs is 200, and early stopping is applied with patience equal to 15. All experiments use a fixed random seed equal to 42. Stability experiments are repeated over five random seeds $\{0,1,2,3,4\}$.

\subsection{Feature Pre-processing for Discovery}
For clustering and incremental discovery, we use the concatenation of the projected I-JEPA embedding and its FSD. Since the two embeddings have different dimensions, we apply standardization and PCA. Both branches are reduced to 64 dimensions and then concatenated, resulting in a final 128-dimensional representation.

\subsection{OOD Rejection Setup}
Open-set rejection is performed using the energy score computed from the classifier logits with temperature T=1. The rejection threshold is selected on the validation set by fixing the false positive rate on known generator samples to 5\%. No samples from unseen generators are used for calibration.
Softmax-based and Generalized Entropy based rejection baselines are evaluated under the same conditions.

\subsection{Clustering Setup}
For clustering rejected samples, UMAP~\cite{mcinnes2020umap} is used as a pre-processing step with 32 output dimensions, 15 neighbors, and a minimum distance of 0.3. Here UMAP is fitted on the pool of energy-rejected samples before HDBSCAN. 
In the offline setting, all rejected samples are available at the same time and are analyzed together. This differs from the incremental setting, where rejected samples arrive progressively and are processed without refitting UMAP.
HDBSCAN is then applied with minimum cluster size equal to 90, minimum samples equal to 15, and epsilon equal to 0.5.
UMAP is also used to produce two-dimensional visualizations of the embedding space. 

\subsection{Incremental Discovery Setup}
The initial reliable space is obtained using HDBSCAN with minimum cluster size equal to 90, minimum samples equal to 15, and epsilon equal to 0.6. For this initialization stage, UMAP is used only to obtain the initial clusters, with 32 output dimensions, 20 neighbors, and a minimum distance of 0.3. For the following incremental stage, UMAP is not performed.
Mahalanobis distance is used to compare each rejected sample with discovered unknown clusters. The matching radius of each reliable cluster is computed using the 0.95 quantile of the cluster distances and a margin equal to 0.95. 
The buffer is processed when it reaches 700 samples. Buffer clustering is performed with HDBSCAN using minimum cluster size equal to 40, minimum samples equal to 5, and epsilon equal to 0.6. 
The buffer size, the minimum cluster support, the cohesion threshold, and the maximum number of buffering attempts were selected through a grid search on the initial reliable space, maximizing the purity of the promoted clusters while keeping their number close to the initialized generators. The resulting configuration promotes a candidate cluster only when it is sufficiently populated (at least 300 samples) and compact (cohesion below 26), discarding samples still unassigned after 4 buffer-clustering attempts.

\subsection{In-the-Wild Setup}
In this experiment, we do not assume prior access to the external test data. Therefore, the embeddings pre-processing is adaptive instead of being fitted on the whole test set. It is initialized on WILD samples and later updated with samples that enter the buffer during the stream. The minimum promotion size is set to 200 samples since the external stream contains fewer samples.

\section{Experiments}
\subsection{Known-Generator Attribution and Open-Set Rejection}

\begin{figure}[!h]
\centering
\includegraphics[width=3.0in]{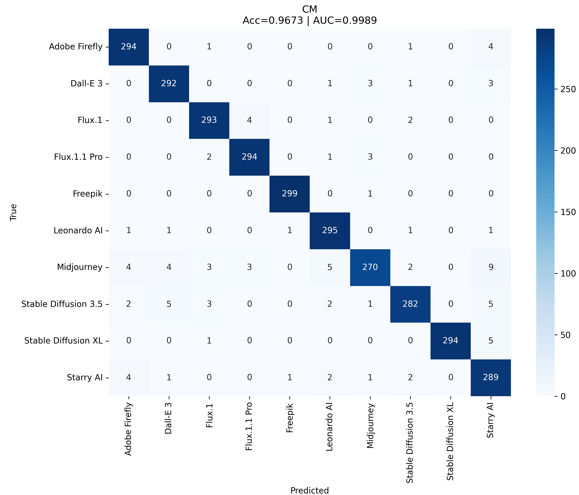}
\caption{Confusion matrix for closed-set attribution of known generators.}
\label{fig_1}
\end{figure}

We first evaluate the attribution of samples produced by the known generators. The confusion matrix in Fig.~\ref{fig_1} shows an almost diagonal structure, and the model achieves an accuracy of 96.73\%, confirming that the learned representation separates the known sources effectively. A detailed comparison with CLIP and DINOv3, including the motivation for selecting I-JEPA as the default backbone, is provided in Section~\ref{sec:backbone_discovery_ablation}.

We then evaluate the model in an open-set attribution setting, where the test samples come from both ID and OOD generators. 
Table~\ref{tab:ood_rejection} compares softmax-based rejection, generalized entropy rejection, and energy-based rejection. Among the evaluated scores, energy-based rejection provides the best balance between the correct acceptance of known samples and the rejection of unknown generators, achieving the highest balanced accuracy of 71.25\%, while retaining a TNR above 94\%. 

To complement the binary rejection analysis, Table~\ref{tab:oscr_results} reports OSCR-based metrics on the WILD closed-set and open-set test splits. All three scores achieve comparable AUOSCR (within $0.002$), indicating similar overall open-set behavior. Energy obtains a slightly lower CCR@FPR5, meaning that, under a strict false-acceptance constraint, it rejects more samples overall, including some correctly classified ID samples. 

Since our pipeline prioritizes routing unknown samples to the discovery stage, we retain energy as the default rejection criterion because it achieves the highest TPR and balanced accuracy.

\begin{table}[!t]
\caption{Open-Set Rejection Results\label{tab:ood_rejection}}
\centering
\begin{tabular}{lcccc}
\hline
Score & BAcc & TNR & TPR & AUROC \\
\hline
Softmax~\cite{hendrycks2016baseline} & 69.38\% & 94.70\% & 44.06\% & 88.53\% \\
Generalized Entropy~\cite{liu2023gen} & 70.09\% & \textbf{94.80\%} & 45.38\% & \textbf{88.83\%} \\
Energy~\cite{liu2020energy} & \textbf{71.25\%} & 94.53\% & \textbf{47.97\%} & 88.63\% \\
\hline
\end{tabular}
\end{table}

\begin{table}[!t]
\caption{End-to-end open-set attribution performance on WILD.}
\label{tab:oscr_results}
\centering
\begin{tabular}{lcc}
\hline
Score & AUOSCR$\uparrow$ & CCR@FPR5$\uparrow$ \\
\hline
Softmax~\cite{hendrycks2016baseline} & 0.8723 & \textbf{0.5917} \\
Generalized Entropy~\cite{liu2023gen} & \textbf{0.8735} & 0.5880 \\
Energy~\cite{liu2020energy} & 0.8715 & 0.5513 \\
\hline
\end{tabular}
\end{table}

\subsection{Clustering of Unknown Generators}
\begin{figure*}[!t]
\centering
\subfloat[]{\includegraphics[width=2.5in]{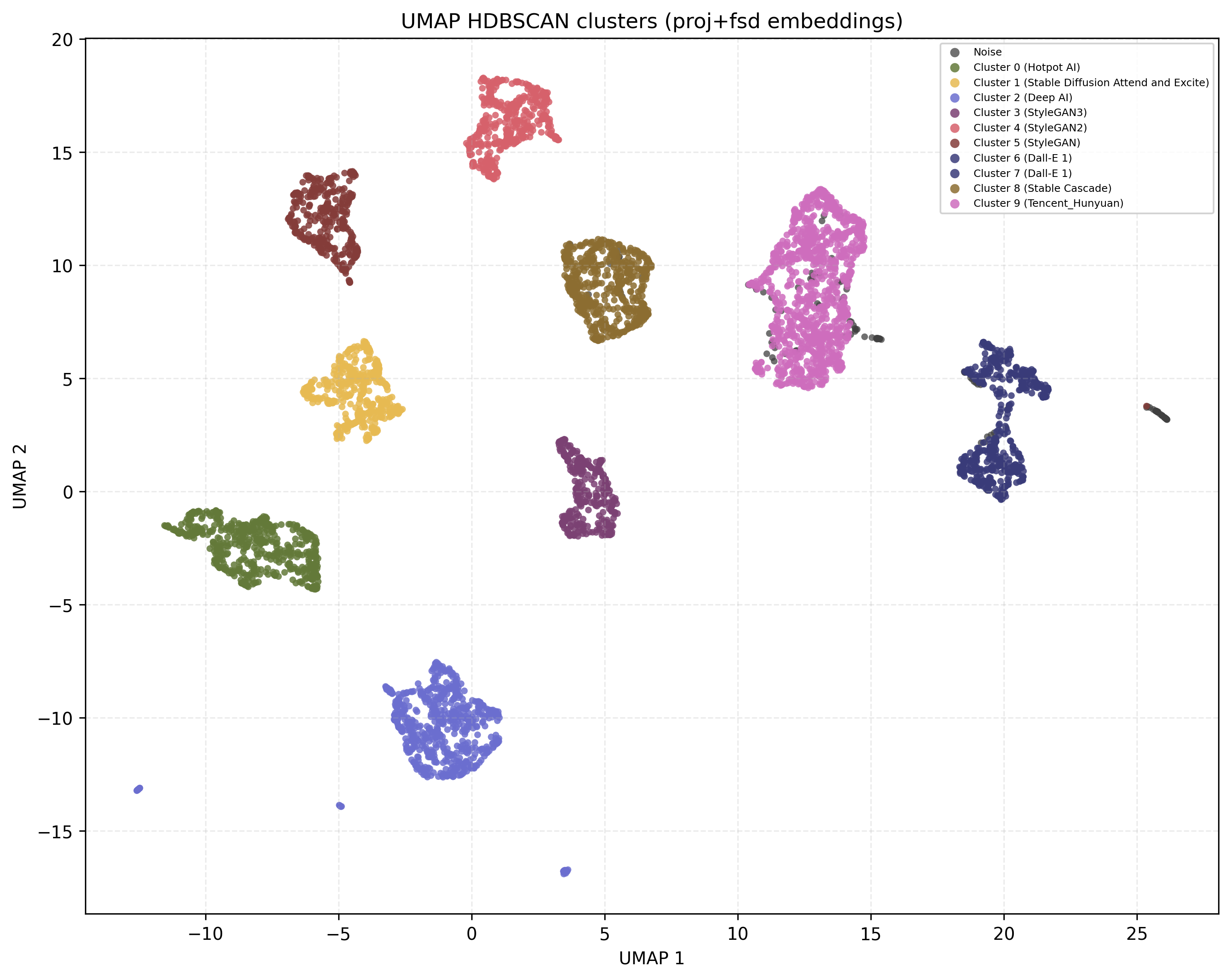}%
\label{fig:energy_cluster_umap}}
\hfil
\subfloat[]{\includegraphics[width=2.5in]{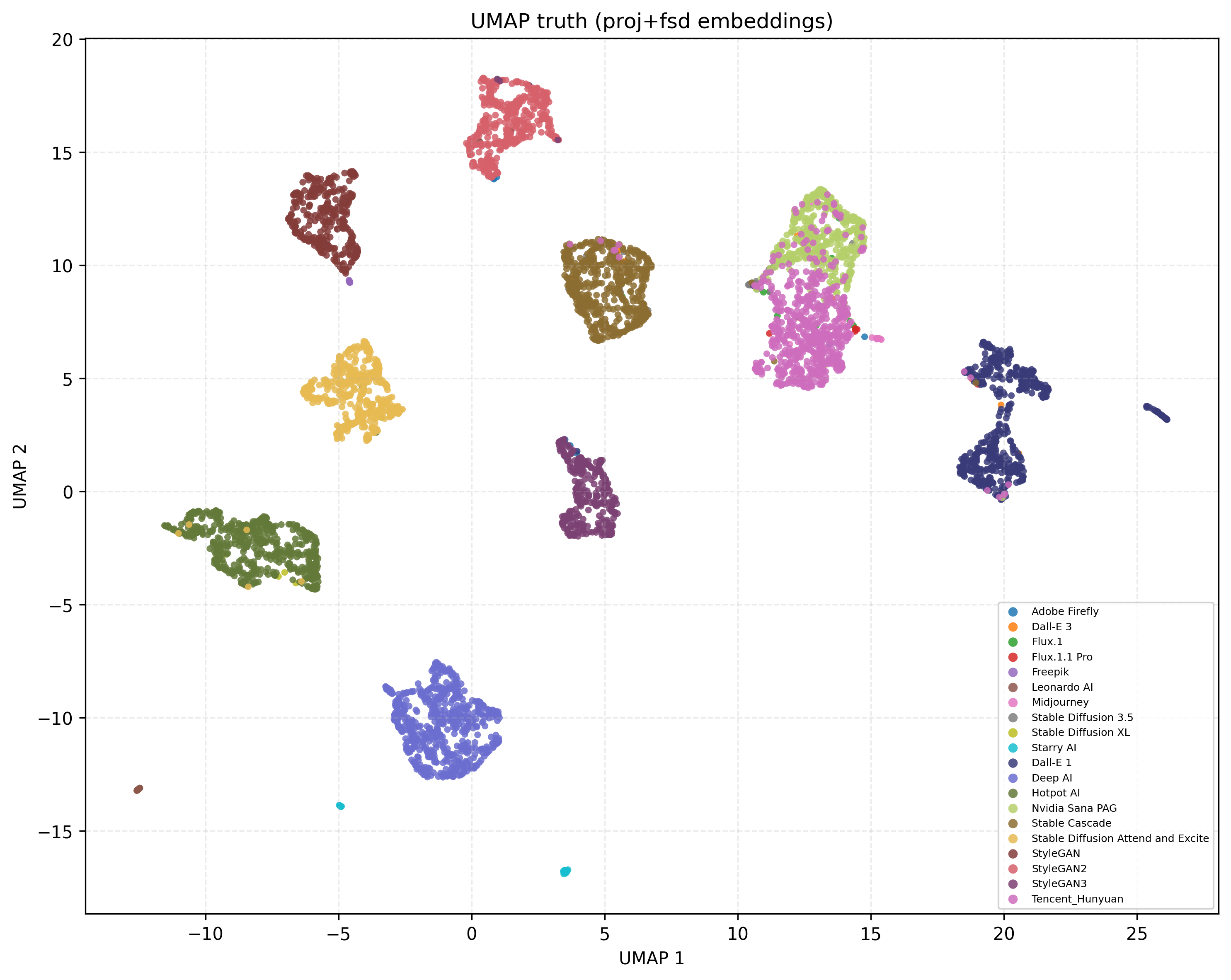}%
\label{fig:energy_truth_umap}}
\caption{UMAP visualization of the clustering results on energy-rejected samples. (a) HDBSCAN cluster assignments (labelled by dominant generator). (b) Ground-truth generator labels. The comparison shows that most generators form compact groups, while the main errors are the split of DALL-E 1 (dark blue) into two separate groups and the overlap between Nvidia Sana PAG (light green) and Tencent Hunyuan (pink).}
\label{fig:energy_umap_comparison}
\end{figure*}

We evaluate the ability of \textit{Face-Trace} to organize rejected unknown samples into generator-specific clusters. In this setting, our method discovers 10 clusters, obtaining an ARI of 0.81, an NMI of 0.90, and an overall purity of 87.74\%. The main errors are the split of DALL-E 1 into two separate high-purity clusters and the merging of Nvidia Sana PAG and Tencent Hunyuan into the same cluster (Fig.~\ref{fig:energy_umap_comparison}).

We further compare the proposed discovery stage with state-of-the-art methods adapted to the WILD benchmark. The comparison includes transductive and non-transductive methods, and the results are reported in Table~\ref{tab:baseline_comparison}. For all baselines, we use the same WILD split adopted in our experimental protocol.

We also evaluate unknown-generator clustering on the post-processed WILD open-set split. As shown in Table~\ref{tab:postprocessing_robustness}, transformations reduce ARI, NMI, and purity because they modify the forensic traces used to distinguish different generators. However, the estimated number of clusters remains close to the ten underlying sources. This suggests that part of the source traces are preserved even when post-processing weakens the separation between the generators.

Overall, the results show that the proposed discovery stage can effectively identify unknown-generator groups under the original data distribution and preserves part of the source structure under increasingly complex post-processing.

\begin{table*}[!t]
\caption{Comparison with state-of-the-art methods on WILD. UDT indicates whether unlabeled target data are available during training; $K$ known indicates whether the number of classes is provided in advance; P/T denotes the predicted and true numbers of novel clusters.}

\label{tab:baseline_comparison}
\centering
\renewcommand{\arraystretch}{1.15}
\setlength{\tabcolsep}{3.5pt}
\begin{tabular}{lcccccc}
\hline
Method & UDT & $K$ known & NMI$\uparrow$ & ARI$\uparrow$ & Purity$\uparrow$ & Novel Clusters (P/T) \\

\hline
SimGCD~\cite{wen2023parametric} 
& Yes 
& Yes 
& 0.82 
& 0.75 
& 81.17\% 
& 14/10 \\

ProtoGCD~\cite{ma2025protogcd} 
& Yes 
& Yes 
& 0.77 
& 0.68 
& 76.18\% 
& 10/10 \\

OWDFA-CAL (w/o KU)~\cite{zheng2025openworld}
& Yes 
& No 
& 0.61 
& 0.44 
& 49.11\% 
& 14/10 \\

OCD~\cite{du2023onthefly}  
& No 
& No 
& 0.32 
& 0.14 
& 41.88\% 
& 214/10 \\
\hline
Face-Trace (Ours) 
& No 
& No 
& \textbf{0.90}
& \textbf{0.81} 
& \textbf{87.74\% }
& 10/10 \\
\hline
\end{tabular}

\vspace{0.3em}
\end{table*}

\begin{table}[!t]
\caption{Robustness of WILD unknown-generator clustering to post-processing. Clustering is performed on all open-set samples.}
\label{tab:postprocessing_robustness}
\centering
\renewcommand{\arraystretch}{1.15}
\begin{tabular}{lcccc}
\hline
Setting & Clusters  & ARI$\uparrow$ & NMI$\uparrow$ & Purity$\uparrow$ \\
\hline
Plain & 9  & \textbf{0.88} & \textbf{0.94} & \textbf{89.20\%} \\
1 Step & 14  & 0.40 & 0.55 & 47.16\% \\
2 Steps & 10  & 0.28 & 0.44 & 42.20\% \\
3 Steps & 9  & 0.15 & 0.26 & 28.83\% \\
\hline
\end{tabular}
\end{table}

\subsection{Incremental Clustering of Unknown Generators}
\begin{figure*}[!t]
\centering
\subfloat[]{\includegraphics[width=2.5in]{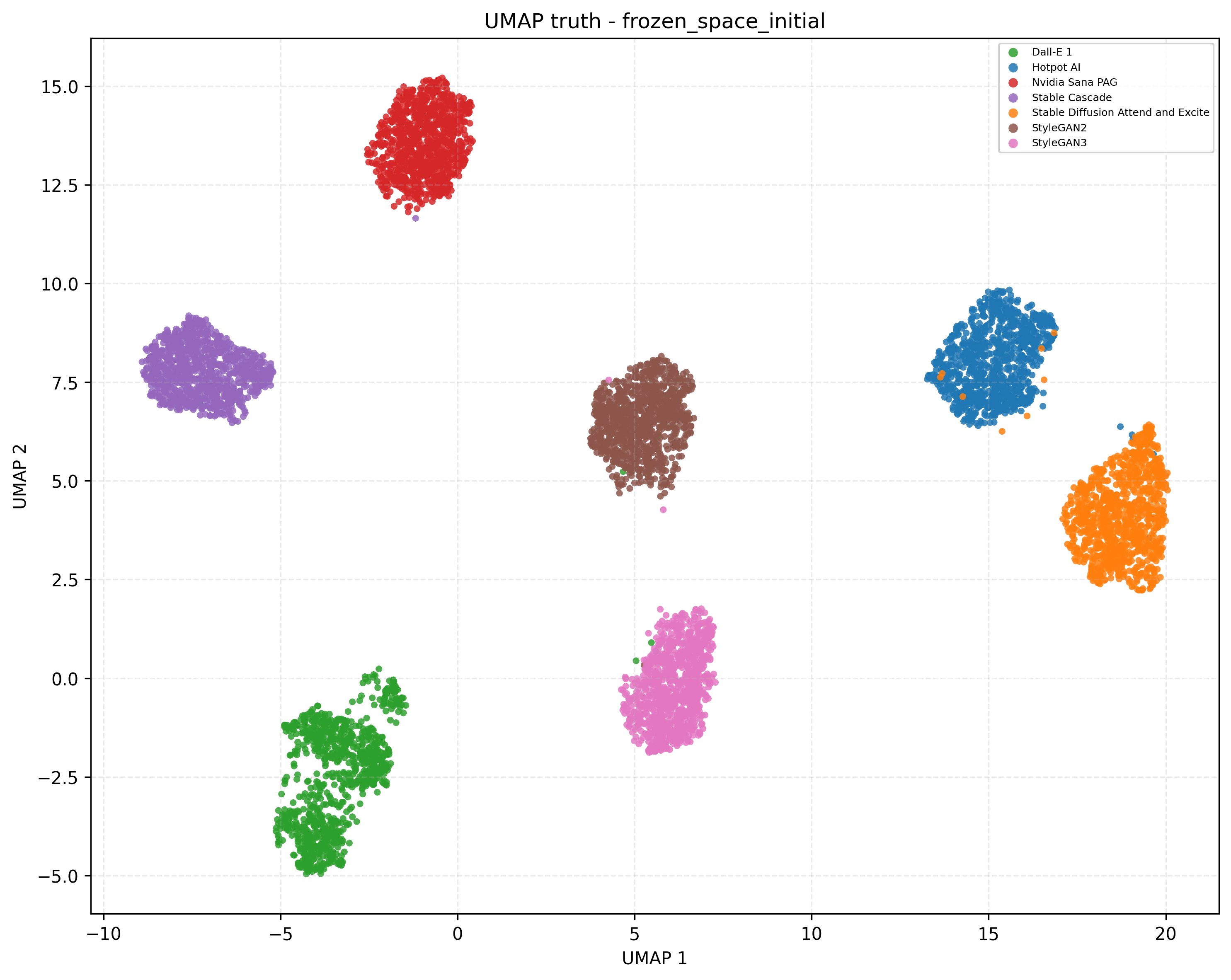}%
\label{fig:incremental_umap_initial}}
\hfil
\subfloat[]{\includegraphics[width=2.5in]{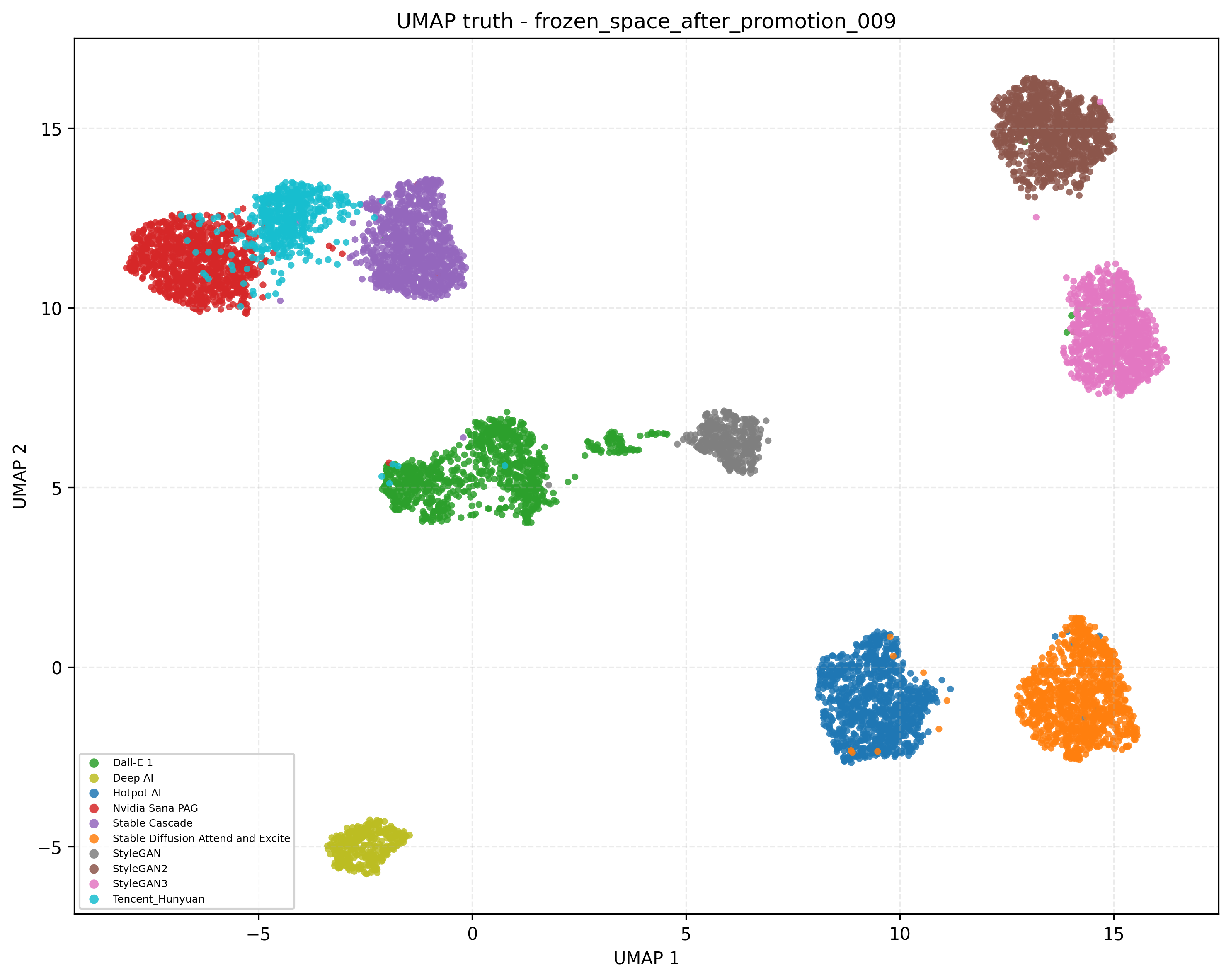}%
\label{fig:incremental_umap_after}}
\caption{UMAP visualization of the reliable unknown space before and after cluster promotion. Points are colored using ground-truth generator labels. (a) Initialized reliable unknown space. (b) Reliable unknown space after promoting new clusters.}
\label{fig:incremental_umap_before_after}
\end{figure*}

For the incremental experiments, the reliable unknown space is initialized using a subset of 7 WILD open-set generators. For each initialized generator, 50 samples are kept aside to test the matching ability of the system with already discovered unknown generators. The 3 remaining generators are treated as novel generators.
The initialized space contains 7 reliable clusters, while the three remaining WILD open-set generators are processed as novel sources to evaluate whether the system can discover and promote new unknown clusters. Table~\ref{tab:incremental_results} reports the results by separating samples from already discovered generators (Known OOD), samples from unknown generators (Novel OOD), and falsely rejected ID samples (Rejected ID). The system directly matches 80\% of the known OOD samples, while all three novel generators are discovered and promoted to the reliable space, with a final generator accuracy of 95.82\%. The rejected ID samples mostly remain unassigned, suggesting that the incremental stage limits the impact of ID contamination. The final reliable space reaches an overall purity of 99.23\%, only slightly lower than the initialized space purity of 99.51\%. The promoted clusters also remain highly pure: 94.95\% for StyleGAN, 97.34\% for Deep AI, and 99.79\% for Tencent Hunyuan. This confirms that the system can extend the unknown space over time (Fig.~\ref{fig:incremental_umap_before_after}).

\begin{table}[!t]
\caption{Incremental Discovery Results\label{tab:incremental_results}}
\centering
\begin{tabular}{lcccc}
\hline
Sample & DMR (\%) & FAR (\%) & SBR (\%) & Accuracy (\%) \\
\hline
Known OOD   & 80.00 & 98.06 &   9.68 & 81.58 \\
Novel OOD   & 11.59 & 84.50 &  17.53 & 95.82 \\
Rejected ID &  5.49 & 5.49  & 100.00 & --    \\
\hline
All         & 16.63 & 78.80 & 25.42  & 93.81 \\
\hline
\end{tabular}
\end{table}

To verify that the incremental results are not specific to the main generator partition, we repeat the experiment using four different combinations of initialized and novel WILD generators. The generator composition of each split is reported in Table~\ref{tab:incremental_generator_splits}. For each configuration, we report the number of reliable clusters obtained during initialization, the number of novel clusters added during the stream, ARI and NMI on the promoted novel clusters, and the purity of the final reliable unknown space. As shown in Table~\ref{tab:incremental_ablation_combinations}, the system recovers all novel generators in two configurations and underestimates their number by only one cluster in the remaining two. The promoted clusters maintain strong agreement with the true generator labels, with ARI between 0.71 and 1.00 and NMI between 0.78 and 1.00. Moreover, final-space purity remains above 97\% in all configurations, showing that the incremental discovery process is robust to different choices of initialized and novel generators.

\begin{table*}[!t]
\caption{Generator combinations used in the incremental split ablation.}
\label{tab:incremental_generator_splits}
\centering
\renewcommand{\arraystretch}{1.15}

\begin{tabular}{p{1.2cm} p{7.0cm} p{6.0cm}}
\hline
Split & Initialized generators & Novel generators \\
\hline
COMB1 & DALL-E 1, StyleGAN2, StyleGAN3, Hotpot AI, Nvidia Sana PAG, Stable Cascade, Stable Diffusion Attend and Excite & StyleGAN, Deep AI, Tencent Hunyuan \\
COMB2 & Tencent Hunyuan, StyleGAN2, StyleGAN3, Deep AI, StyleGAN, DALL-E 1, Stable Cascade & Nvidia Sana PAG, Hotpot AI, Stable Diffusion Attend and Excite \\
COMB3 & Stable Diffusion Attend and Excite, StyleGAN2, Hotpot AI, Deep AI, StyleGAN, DALL-E 1, Nvidia Sana PAG & Stable Cascade, StyleGAN3, Tencent Hunyuan \\
COMB4 & Stable Diffusion Attend and Excite, StyleGAN2, Deep AI, StyleGAN, Nvidia Sana PAG, DALL-E 1 & Stable Cascade, StyleGAN3, Tencent Hunyuan, Hotpot AI \\
\hline
\end{tabular}
\end{table*}

\begin{table*}[!t]
\caption{Results on different WILD generator splits for incremental discovery.}
\label{tab:incremental_ablation_combinations}
\centering
\renewcommand{\arraystretch}{1.15}
\begin{tabular}{lccccccc}
\hline
Split & Real init. Gen. & Init. Gen. & Real Novel Gen. & Novel Gen. & ARI$\uparrow$ & NMI$\uparrow$ & Purity$\uparrow$ \\
\hline
COMB1 & 7 & 7 & 3 & 3 & 1.0 & 1.0 & 99.23\% \\
COMB2 & 7 & 7 & 3 & 3 & 0.71 & 0.78 & 97.19\%  \\
COMB3 & 7 & 7 & 3 & 2 & 1.0 & 1.0 & 98.31\% \\
COMB4 & 6 & 6 & 4 & 3 & 0.99 & 0.98 & 99.57\% \\
\hline
\end{tabular}
\end{table*}

\subsection{In-the-Wild Incremental Clustering}
We further evaluate the incremental discovery stage in a cross-dataset setting. While the previous experiments are conducted entirely on WILD, this experiment introduces external samples to evaluate whether the unknown space built on WILD can handle unseen generators coming from different data distributions.
The space of discovered generators is initialized using samples from the WILD open-set split, while the external data is organized as follows: samples from a generator already present in the closed-set split (Known ID), samples from a generator already present in the initialized unknown space (Known OOD), and samples from two generators that should be discovered by the system (Novel OOD). 
The results show that the energy rejection module keeps most ID samples out of the discovery stage, while 45.33\% of the OOD samples are correctly rejected. DALL-E 3 samples from the external dataset are mostly attributed to the corresponding WILD closed-set generator, suggesting that ID attribution remains robust. StyleGAN2 samples from an external dataset are partially matched to the previously discovered StyleGAN2 cluster, indicating that cross-dataset matching of already discovered unknown generators is possible, although more sensitive.
Starting from 11 initial reliable clusters, the system ends with 13 reliable clusters, which means that two novel clusters are promoted during the stream. The promoted novel clusters achieve an ARI of 0.33, an NMI of 0.40, and a purity of 78.94\% due to the contamination of AttGAN into the Latent Diffusion cluster.

Overall, these results (Tab.~\ref{tab:itw_incremental_results}) show that the proposed incremental pipeline can operate in-the-wild.

\begin{table}[!h]
\caption{In-the-wild full pipeline results. "Known ID/OOD" aggregates external samples from a generator already present in the closed-set split and from a generator already present in the initialized unknown space.}
\label{tab:itw_incremental_results}
\centering
\begin{tabular}{lcccc}
\hline
Sample & DMR (\%) & FAR (\%) & SBR (\%) & Accuracy (\%) \\
\hline
Known ID/OOD & 94.31 & 94.31 & 100.00 & 83.03 \\
Novel OOD & 51.40 & 99.25 & 1.54 & 39.19  \\
Rejected ID & 8.54 & 8.54 & 100.00 & -- \\
\hline
All & 80.04 & 93.71 & 31.50 & 69.59 \\
\hline
\end{tabular}
\end{table}

\section{Ablation Studies}
\subsection{Stability Across Random Seeds}
We evaluate pipeline stability over five random seeds, using each seed for head training and the stochastic components of clustering and incremental discovery, while keeping dataset splits, backbone weights, and hyperparameters fixed.
The mean and standard deviation results in Tab.~\ref{tab:seed_stability} show limited variability overall. Novel OOD FAR has the highest variance ($88.29 \pm 5.96\%$), whereas assignment accuracy remains high ($94.78 \pm 1.88\%$), indicating that uncertain samples are more often left unassigned than incorrectly promoted.

\begin{table}[!t]
\caption{Stability analysis over five random seeds $\{0,1,2,3,4\}$. Offline discovery is evaluated on all open-set samples. Results are reported as mean $\pm$ standard deviation.}
\label{tab:seed_stability}
\centering
\renewcommand{\arraystretch}{1.15}
\setlength{\tabcolsep}{5pt}
\begin{tabular}{lcc}
\hline
Experiment & Metric$\uparrow$ & Result \\
\hline
Closed-set attribution 
& Accuracy 
& $96.83 \pm 0.27\%$ \\
\hline
Offline discovery 
& ARI 
& $0.8746 \pm 0.0171$ \\

Offline discovery 
& NMI 
& $0.9105 \pm 0.0121$ \\

Offline discovery 
& Purity 
& $89.82 \pm 1.41\%$ \\
\hline
Incremental discovery 
& Known OOD FAR 
& $99.48 \pm 0.24\%$ \\

Incremental discovery 
& Novel OOD FAR
& $88.29 \pm 5.96\%$ \\

Incremental discovery 
& Novel OOD accuracy 
& $94.78 \pm 1.88\%$ \\
\hline
\end{tabular}
\end{table}

\subsection{Backbone Representation for Attribution and Discovery}
\label{sec:backbone_discovery_ablation}

To better assess the contribution of the visual representation, we compare I-JEPA with other frozen visual backbones, namely CLIP and DINOv3. 
For a fair comparison, all methods use the same lightweight attribution head and are evaluated on the same closed-set test split. 

\begin{table}[!h]
\centering
\caption{Closed-set attribution performance with different frozen visual backbones.}
\label{tab:closed_set_backbone_comparison}
\begin{tabular}{lcc}
\hline
Backbone & Accuracy & AUC \\
\hline
CLIP~\cite{radford2021clip}   & \textbf{98.33}\% & \textbf{99.96}\% \\
DINOv3~\cite{simeoni2025dinov3} & 97.60\% & 99.95\% \\
I-JEPA~\cite{assran2023ijepa} & 96.73\% & 99.89\% \\
\hline
\end{tabular}
\end{table}

As shown in Table~\ref{tab:closed_set_backbone_comparison}, all three backbones provide strong closed-set attribution performance. Although I-JEPA achieves a lower accuracy than CLIP and DINOv3, backbone selection is based on the overall pipeline rather than on closed-set attribution alone. 
We therefore extend the comparison to the discovery stage, where the representation must also preserve source-specific differences among unseen generators. For all backbones, the projected visual representation is fused with FSD features, and the same UMAP and HDBSCAN configuration is used.
Table~\ref{tab:backbone_fsd_ablation} shows a different trend from the closed-set results. CLIP + FSD obtains lower ARI and purity, whereas DINOv3 + FSD and I-JEPA + FSD provide substantially stronger cluster separation. Their NMI values are almost identical, but I-JEPA + FSD achieves the highest ARI and purity. We therefore adopt I-JEPA as the default backbone.

\begin{table}[!t]
\caption{Backbone ablation for WILD unknown-generator clustering on all open-set samples.}
\label{tab:backbone_fsd_ablation}
\centering
\renewcommand{\arraystretch}{1.15}
\setlength{\tabcolsep}{5pt}
\begin{tabular}{lcccc}
\hline
Embedding & Clusters & ARI$\uparrow$ & NMI$\uparrow$ & Purity$\uparrow$ \\
\hline
CLIP + FSD & 8 & 0.72 & 0.89 & 78.88\% \\
DINOv3 + FSD & 9 & 0.87 & \textbf{0.94} & 88.75\% \\
I-JEPA + FSD & 9 & \textbf{0.88} & 0.94 & \textbf{89.20\%} \\
\hline
\end{tabular}
\end{table}

\subsection{Clustering Method Comparison}
Table~\ref{tab:clustering_algorithm} compares different clustering strategies on the complete OOD set. Among the methods that do not require the number of generators in advance, HDBSCAN achieves the best ARI and NMI. GMM reaches a higher purity, but only by producing roughly twice the number of underlying sources, since purity grows mechanically with over-segmentation. Fixed-$K$ results are reported only as references, as they assume knowledge unavailable in an open-set setting.

\begin{table}[!t]
\caption{Clustering algorithm ablation on all open-set samples, including noise-labelled samples. Fixed-$K$ results are reported as references.}
\label{tab:clustering_algorithm}
\centering
\renewcommand{\arraystretch}{1.15}
\setlength{\tabcolsep}{4pt}
\begin{tabular}{llcccc}
\hline
Method & $K$ selection & Clusters & ARI$\uparrow$ & NMI$\uparrow$ & Purity$\uparrow$ \\
\hline
K-Means 
& Silhouette 
& 7 
& 0.6633 
& 0.8711 
& 69.49\% \\

K-Means 
& Elbow 
& 5 
& 0.5618 
& 0.7988 
& 49.92\% \\

K-Means 
& Fixed $K=10$ 
& 10 
& 0.9482 
& 0.9537
& 97.52\% \\

\hline
GMM 
& BIC 
& 19 
& 0.7496 
& 0.8658 
& 97.40\% \\

GMM 
& AIC 
& 20 
& 0.7136 
& 0.8550 
& 97.40\% \\

GMM 
& Fixed $K=10$ 
& 10 
& 0.9443
& 0.9529 
& 97.26\% \\

\hline
HDBSCAN 
& Automatic 
& 9 
& 0.8684 
& 0.9314 
& 88.99\% \\

\hline
\end{tabular}
\end{table}

\subsection{Embedding Type}
We evaluate the impact of the embedding representation on WILD unknown generator clustering. We test different types of representation, including embeddings from projection head trained with Supervised Contrastive Loss (SupCon) and Cross-Entropy. While Cross-Entropy mainly optimizes the decision boundary between known classes, SupCon learning forces samples from the same class to be close in the representation space and samples from different classes to be farther apart~\cite{khosla2020supervised}. 

For this ablation we use all samples from the WILD open-set split without the intermediate energy rejection step. All embeddings are evaluated using the same HDBSCAN parameters. Table~\ref{tab:embedding_ablation} shows that the frozen I-JEPA representation is not sufficient to obtain a separated generator space. 
Interestingly, projecting I-JEPA features with a CE-trained head alone further degrades the clustering structure, while adding a supervised contrastive term partially recovers the structure.
FSD features provide strong clustering performance, while the best results are obtained by fusing FSD with I-JEPA embeddings. In particular, the FSD + I-JEPA representation trained with CE achieves the highest average purity, showing that this combination produces the most discriminative space for unknown generator clustering.

\begin{table}[!t]
\caption{Ablation study on the embedding representation for WILD unknown-generator clustering. Clustering is performed on all open-set samples.}
\label{tab:embedding_ablation}
\centering
\renewcommand{\arraystretch}{1.15}
\begin{tabular}{lcccc}
\hline
Embedding & Clusters & ARI$\uparrow$ & NMI$\uparrow$ & Purity$\uparrow$  \\
\hline
I-JEPA raw & 8 & 0.18 & 0.37 & 36.00\% \\
I-JEPA CE & 2 & 0.00 & 0.01 & 14.53\% \\
I-JEPA CE+SupCon & 3 & 0.23 & 0.49 & 29.77\% \\
FSD & 7 & 0.66 & 0.85 & 70.15\% \\
FSD + I-JEPA CE & 9 & \textbf{0.88} & \textbf{0.94} & \textbf{89.20\%} \\
FSD + I-JEPA CE+SupCon & 9 & 0.87 & 0.94 & 88.74\% \\
\hline
\end{tabular}
\end{table}

\begin{table}[!h]
\caption{Ablation study on HDBSCAN parameters for WILD generators clustering, including noise samples. Clustering is performed on energy-rejected samples.}
\label{tab:hdbscan_ablation}
\centering
\renewcommand{\arraystretch}{1.15}
\resizebox{.5\textwidth}{!}{%
\begin{tabular}{ccccccc}
\hline
min\_clust\_size & min\_sampl & $\epsilon$ & Clust. & ARI$\uparrow$ & NMI$\uparrow$ & Purity$\uparrow$ \\
\hline
40 & 5 & 0.7 & 6 & 0.46 & 0.77 & 60.10\% \\
60 & 10 & 0.6 & 8 & 0.65 & 0.85 & 76.63\% \\
90 & 15 & 0.5 & \textbf{10} & 0.79 & \textbf{0.88} & 86.33\% \\
120 & 20 & 0.4 & 11 & \textbf{0.80} & 0.86 & \textbf{91.73\%} \\
\hline
\end{tabular}
}%
\end{table}

\subsection{HDBSCAN Parameters}
Table~\ref{tab:hdbscan_ablation} reports the results of WILD unknown generator clustering with different HDBSCAN configurations. Here metrics are computed including noise-labelled samples, in order to account for the fraction of data discarded by each configuration.
Less conservative settings produce fewer and less pure clusters, while increasing the parameters improves purity and leads to a number of clusters closer to the expected number of WILD open-set generators. 
The configuration with min\_cluster\_size = 90, min\_samples = 15, and $\epsilon = 0.5$ achieves the best overall balance across ARI, NMI, purity, and number of discovered clusters. For this reason, it is adopted in the conducted experiments.

\section{Conclusion}
In this work, we introduced \textit{Face-Trace}, a framework that extends open-set synthetic face attribution beyond the simple rejection of unknown generators by organizing rejected samples into generator-specific groups. Its main novelty lies in the non-transductive and incremental formulation: the classifier is trained only on known generators, while unknown generator samples become available only at test time. The attribution and rejection modules remain fixed after training, whereas the reliable unknown-generator space is progressively expanded as new samples arrive, avoiding classifier retraining whenever a new source is discovered.

The closed-set results show that the learned representation remains highly discriminative for known-generator attribution, whereas the discovery experiments further demonstrate that the fused I-JEPA and FSD representation produces coherent unknown-generator clusters.
The incremental experiments show that the reliable unknown-generator space can be extended while preserving high cluster purity and accurate assignments. The results across different generator splits indicate that this behavior is not limited to a single initialization. Moreover, the cross-dataset evaluation shows that the same space can be reused to process samples from known generators, previously discovered unknown sources, and completely unseen generators coming from external datasets. Post-processing remains the most challenging condition, however the experiments show that part of the unknown-source structure is still preserved. 

Future work will focus on improving robustness to post-processing, extending the framework beyond facial images to videos and other media, and evaluating longer incremental streams with different buffer and promotion settings.

\end{document}